# Developing Courses With HoloRena, a Framework for Scenario- and Game Based E-Learning Environments

Laszlo Juracz

Institute for Software Integrated Systems, Vanderbilt University, Nashville, U.S.A.
`laszlo.juracz@vanderbilt.edu`

## ABSTRACT

*However utilizing rich, interactive solutions can make learning more effective and attractive, scenario- and game-based educational resources on the web are not widely used. Creating these applications is a complex, expensive and challenging process. Development frameworks and authoring tools hardly support reusable components, teamwork and learning management system-independent courseware architecture. In this article we initiate the concept of a low-level, thick-client solution addressing these problems. With some example applications we try to demonstrate, how a framework, based on this concept can be useful for developing scenario- and game-based e-learning environments.*

## 1. INTRODUCTION

Learning by doing [1] and other new instructional paradigms benefited from the combined availability of multimedia, interactivity and adaptive learning technologies. These paradigms spawned learning strategies, such as online scenario-based learning (SBL) [2] and role-play simulation. These strategies enable immersion in real world like problems, and interaction with a rich context thereby increasing usability. Most of the scenario-based learning resources enact content in the form of sequences of interactive slides, while agents guard the progress of the learner through the course taking care of adaptive sequencing, learner profiling and more sophisticated adaptive mechanisms.

Research interest, learners' need for rich, interactive virtual environments and the popularity of computer games brought huge attention to digital game-based learning (DGBL), including game-based e-learning [3]. There is a broad range of DGBL design initiatives from repurposing existing games to education, to transforming existing official educational content into game-like environments [4]. Although, repurposing existing online games is hardly feasible, the vision of a general framework that enables the reuse of existing game components is realizable; it is also necessary in order to achieve flexibility and cost effectiveness. Game-based online courses utilize interactivity, situate the learner in graphically simulated environments, and allow continuous interaction with virtual objects besides possessing phasing or sequencing aspects.

Developing GBL or SBL resources is a teamwork of experts from various fields [5]. Educational content is authored by a group of subject matter experts (SME) and instructional designers. Artists and programmers develop visual, interactive and multimedia elements. Cresting and changing programmed components is usually very expensive and time-consuming.

Production of advanced learning resources is a complex process with feedback loops, this necessitates easy modification of these resources anytime during the development process.

In the online setup educational applications are executed in a browser on the client side. The client maintains a local state of the courseware progress and synchronizes this state with the remote Learning Management System (LMS) on-demand. Unfortunately, standard LMSs only provide limited ways to implement adaptation, student evaluation and storage solutions. Communication with the server can take a significant amount of time; frequent synchronous calls to a remote server can block the client from providing a coherent and smooth user experience. However this smooth "flow" experience is a crucial factor in digital game design and it is the fundamental virtue of any engaging learning environments.

Paradigms, like learner-centred design [6], are based on the affordances of adaptive web-based learning. Adaptation enables individualized learning content, intelligent analysis of student's solutions, collaboration support and adaptive presentation and navigation [7]. Good SBL and DGBL educational resources should employ guards which take care of keeping the learner in the flow while immersed in the learning activity [8]. The implementation of these guards, and a sophisticated game engine implies real-time, continuous monitoring and evaluation of user interaction in the learning environment.

The combination of methodologies, such as the object oriented construction approach [9, 10] and modelling frameworks, like the adaptive hypermedia application model [11] makes the realization of intelligent, customized learning systems feasible. Authoring software, such as the CAPE authoring environment [12], delivery engines, like the eLMS adaptive learning platform [13], make it possible to design, create and disseminate content applying sophisticated learning strategies. Most of these tools [14] also support the creation of SCORM packages, although sometimes with decreased functionality, due to limitations in the SCORM data-storage specification.

Based on experiences from past work, and the requirements for our upcoming courseware development projects we were looking for a framework that enables isolation of the presentation interface from the story content, supports the concept of reusable, skinnable components and facilitates rapid modifications and creation of products.

Although numerous tools fulfil some of these initial requirements, they mostly support the development of static, environment-based, sequenced learning materials; they do not support the concept of dynamic, cross-scene or global objects and LMS-agnostic design.

Eventually we came up with the idea of a thick-client solution, where the educational application runs in the browser and it implements most of the above-mentioned adaptive strategies with the limited use of remote LMS features. In this configuration, modularity and reusability is achieved by developing content specific display components or generic, intelligent modules. A low-level, educational strategy- and courseware design independent framework coordinates the collaboration of these autonomous, self- and context-aware components. Basic sequencing maps provide the skeleton of the courseware progress, but adaptive mechanisms can alter the flow any time it is necessary. This framework also provides ways to enable sequencing intervention, data persistence, content and components loading and access to a locally (client-side, browser-based) shared data-space. It is the concept, which served as the basis of our HoloRena framework [15].

Programming interactions in the client's browser can be implemented in JavaScript; multimedia, audio and video content can be displayed with HTML and CSS features; however developing complex applications in standard HTML targeting a wide set of browsers on different operating systems can still raise compatibility issues. Despite that browsers started to support some aspects of the being developed HTML5 and CSS3 specification, the speed and the smoothness of rendering the visual content can vary significantly [16]. To eliminate compatibility problems, and to be able to provide an outstanding, game-like, highly interactive experience we think, that among all the currently available technologies, Adobe Flash is a good platform for developing educational games on the web.

This paper focuses on the concept of a general framework of SBL and GBL environments on the web and the comprehensive overview of an implementation of such a framework in Adobe Flash/Flex. We also give an impression of the utilization of this framework through the presentation of some practical applications.

## 2. RELATED WORK

Research efforts and commercial solutions dealing with our initial ideas and requirements fall under the mixed discipline of e-learning system engineering and educational game design [17, 4]. Our most important question was how could we implement interactive components that we could use across several, interactive adaptive learning resources.

Reusability is usually addressed through promoting modularity and the use of learning objects (LOs) in frameworks, like MALS [9] and methodologies, like the Sequence Feature Diagrams and Metaprogramming Techniques [10]. These approaches operate on the level of the learning content structure.

Since LOs embody both the visual representation and the specification of the learning content, they are tightly connected with their context in the educational scenario. It puts a limitation on their reusability. This limitation in some cases can be acceptable. For example, if a component of an educational game is not reusable for any other similar educational game project but can be used in several places in the original educational game, the component can still be considered as reusable.

The GLO Maker authoring tool [18] features the concept of generative learning objects (GLOs) [19]. GLOs take the vision of repurposing even higher, they are designed to express the underlying pedagogical design of LOs. GLO Maker is an impressive tool and framework for Flash-based, tutorial-like educational resources. It introduces the concept of flexible visual templates, providing a solution to decouple presentation from the learning content. Unfortunately its sequencing mechanism lacks the support of advanced adaptivity, it does not provide ways to integrate GLOs with an LMS, furthermore the provided visual templates are not extendible enough to implement custom, highly interactive virtual environments.

In [20] Hang and Kramer demonstrate a methodology for developing reusable learning resources through the design of customizable, interactive learning objects implemented as Flash/Flex software components. Although they envision the collaboration of these kind of software components with other, third party applications, their results does not cover any details of a complex framework coordinating the use of customizable, interactive LOs in a complex e-learning resource.

## 3. THE CONCEPT OF HOLORENA

The knowledge of web-based learning systems and the requirements we identified led us to the decision to develop a framework that we can use for our in-house experiments and projects. HoloRena was designed to implement all the features we think a modern web-based educational framework should possess.

### 3.1 Adaptive sequencing

In web-based learning environments the learning content is typically presented as a series of interactive screens. In more interactive forms of e-learning, such as SBL and DGBL these screens represents complex activities. In our framework the sequence of these learning activities is adaptive; it is governed by a set of composition rules that take into consideration the learner's performance, thereby making the experience flow-like for the individual. In the sequencing model of HoloRena, these stationary screens are called *Scenes*.

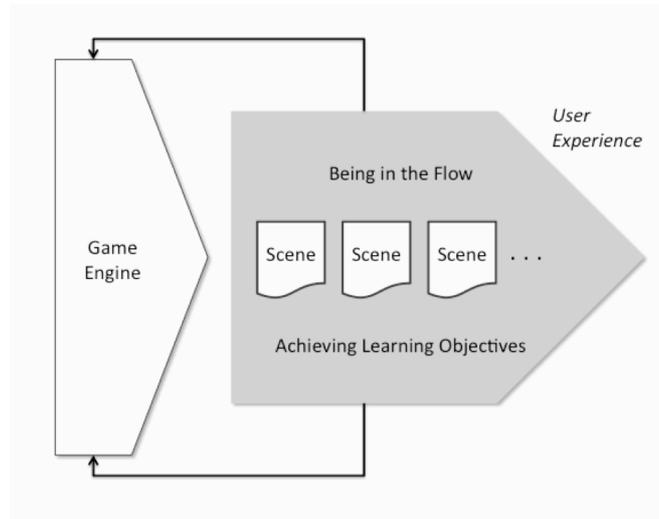

Figure 1. Going with the flow in the learning environment [15]

Sequence segments in the content specification are called *Streams*. A *Stream* can include interactive elements: *Scenes*, non-interactive script elements: *Actions,* and other *Streams*. *Actions* are code-snippets written in a script language that act upon a globally visible data space and thus manipulate sequencing decisions. *Scenes*, *Actions* and *Streams* are collectively referred to as *StoryElements*.

For every HoloRena course, a course flow model should be compiled. This course flow model is a tree-like hierarchy of *StoryElements* starting with a single *Stream*, called the *RootStream*. This structure enables course designers to describe complex instructional patterns.

Execution of the course flow starts with the very first includable *StoryElement* in the *RootStream*. When the execution of this *StoryElement* has ended, the execution engine will find the next includable *StoryElement* according to the current execution mode. The possible execution modes are the following:

- In *REGULAR_MODE* the execution engine takes the next includable *StoryElement* in the current stream. This is the default mode: *StoryElements* are always executed according to their order in the given *Stream*. This mode is also called the downstream mode. It is the only stable mode: after each execution, the execution mode is set back to this mode automatically, if a different mode is not set explicitly. If the end of a *Stream* is reached, the execution engine looks for the next executable element in the parent *Stream*. When the end of the *RootStream* is reached, the execution engine jumps back to the very beginning of the *RootStream*.
- If *STAY_MODE* is set, the execution will try to include the same *StoryElement* again.
- *UPSTREAM_MODE* tells the execution engine to attempt to include the preceding *StoryElement* within the same *Stream*. If there is no previous element in the *Stream*, the execution engine switches to the parent *Stream*, unless it was in the *RootStream*.

- With setting *CANAL_MODE*, the execution engine can be directed to navigate to a *StoryElement* identified by its absolute *path* in the course flow.

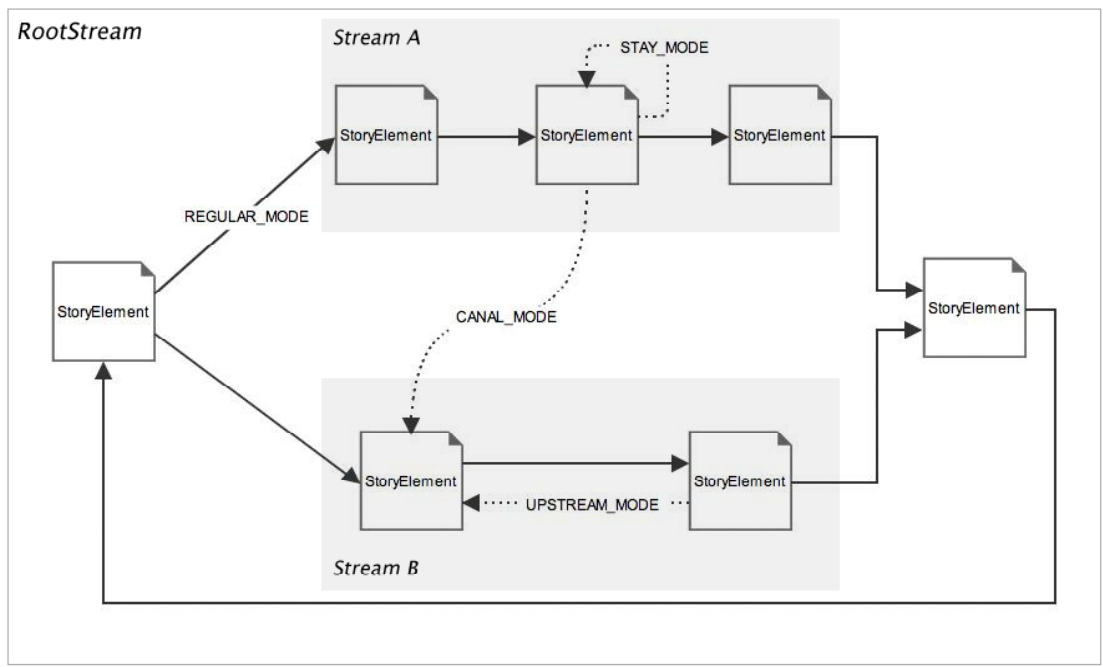

Figure 2. HoloRena execution modes

A *StoryElement* can be included for execution if its *includeIf* attribute in the course flow specification is not set or evaluated as *true*. The *includeIf* attribute is a function written in the script language of *Actions,* a simplified version of ECMAScript. This attribute is a common place to apply adaptive mechanisms; it is generally used for simple branching decisions in the course flow.

During the execution of a *StoryElement* the execution engine is passive, it waits for calls triggering the next execution. The execution engine is automatically triggered whenever a *StoryElement* dispatches the *complete* event:

- An *Action* is completed, when the evaluation of the script block is over.
- *Scenes* can become completed by some internal mechanism. Usually some user interaction is needed to achieve this state.
- When the execution of the last includable *StoryElement* in a *Stream* is over, the Stream becomes completed.

A non-automatic triggering call is the *interrupt*. It is a globally accessible HoloRena service, which terminates the execution of *Scenes* and *Streams* from outside, independently from their current internal state.

*StoryElements* can have the following event handlers specified:

- *onExecute* is executed right before the executer enacts the *StoryElement*
- *onComplete* is called after the *StoryElement* is completed and before the executer steps in action
- *onInterrupt* is executed after the StoryElement was interrupted

Event handlers turn out to be very useful to initialize and set up environments and courseware-state variables related to the actual sequencing.

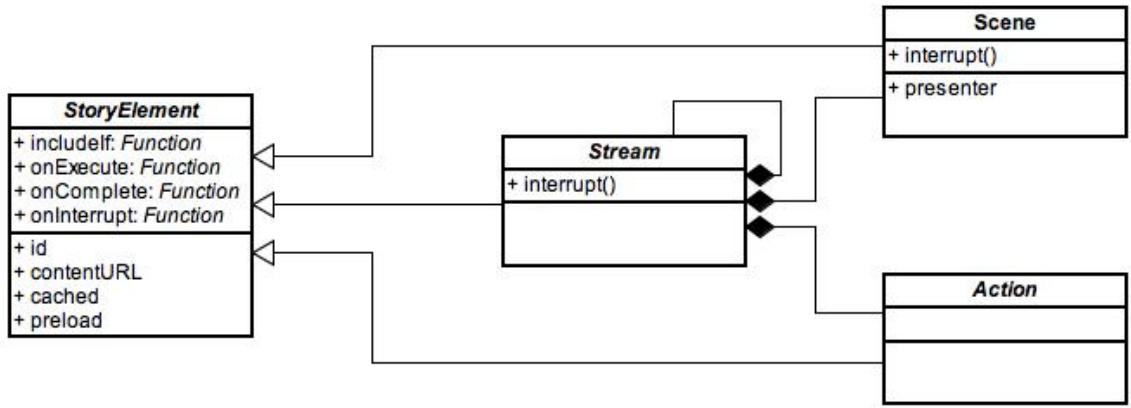

Figure 3. HoloRena *StoryElement* types

## 3.2 Course content structure and segmentation

### 3.2.1 Course flow specification

The HoloRena course flow, and most of the non-visual course content is specified in XML format. The execution engine starts with the first includable *StoryElement* described in the downloaded XML document of the *RootStream*. Content for a *StoryElements* can be embedded in the encapsulating *Stream* XML, or can be specified in an external XML document. Externalized contents will be downloaded on-demand, only when the execution engine is about to execute the given *StoryElement*, unless its *preload* attribute is set to *true*. A *StoryElement* and its content can also be set to *cached*. These features allow the designer to apply the following strategies:

- Unnecessary loading of flow segments can be avoided.
- Planned preloading can be implemented before intense, fast flow sections to avoid glitches in the flow experience.
- Similar flow blocks used at several places in the course flow should be implemented only once and can be included externally at locations where needed.
- Looping or returning flow segments can be cached to avoid extra downloading.

```xml
<streamContent>
    <stream id="init" includeIf="!Global['initializedAlready']" >
        <stream id="dataInit" contentURL="perpsContent/fct/xml/data.xml"/>
        <action id="commonInit" contentURL="perpsContent/main_init.xml"/>
        <action id="courseSpecificInit" contentURL="perpsContent/fct_init.xml"/>
        <action id="couseStateInit" contentURL="perpsContent/course_state_init.xml"/>
    </stream>
    <scene id="disclaimer" presenterType="DHSPresenter" includeIf="!Global['skipLogoAni']"/>
    <scene id="survey" presenterType="PilotSurvey" onComplete="Global['onSurveyComplete']()"/>
    <scene id="map" presenterType="CourseMapPresenter" preload="true"/>
    <stream id="test" onExecute="Global['onTestStart']()" onInterrupt="Global['onTestPaused']()"/>
    <stream id="remediation" onExecute="Global['onRemediationStart']()"/>
</streamContent>
```

Figure 4. Part of a flow content specification

The framework also provides a service to set/overwrite the content of any *StoryElement*, dynamically. This powerful feature can be utilized to alter course flows adaptively.

### 3.2.2 Interactive and visual components

Besides its content specification, each *Scene* has a linked, interactive, visual component, which implements the user interface and renders the content on the screen. These kinds of components in HoloRena are called *Presenters*. *Presenters* can be reusable: typically they are linked to several different *Scenes* in a course.

The framework also supports the concept of cross-scene objects usually taking the responsibility for some part of the visualization (e.g. UI components facilitating navigation, showing course progression). *Gadgets* are not directly related to the execution of *StoryElements*. They can implement asynchronous mechanisms (e.g. timers) independently from the execution engine. *Gadgets* do not necessarily have visual representation, they can be considered as plug-ins.

*Presenters* and *Gadgets* are *SubApplications*. Technically *SubApplications* are compiled Flash or Flex applications. They can be preloaded and cached similarly to *StoryElements*.

### 3.3 RIA architecture

HoloRena is a Rich Internet Application framework [22]. Its core, the main application is the *Player* which implements the execution engine, globally accessible services, global data space, manages the content and resources and implements interfaces towards the LMS.

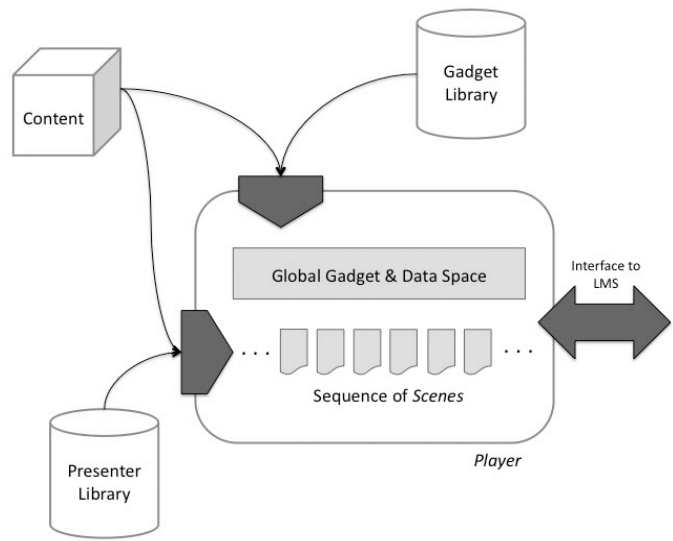

Figure 5. HoloRena architecture

*SubApplications* and the player are implemented as Flash and Flex applications. They provide the basis for developers to extend the traditional Web architecture by moving part of the application data and computing logic to the client-side with the primary aim of providing an autonomous and reactive user interface. The programming of *SubApplications* should accommodate the principles of developing multi-versioned applications in Flex 4 [23].

HoloRena supports standard, SCORM compliant communications and facilitates the integration with eLMS [13] too. The framework also provides a globally accessible journaling service, which should be called with a *string* event-identifier, timestamps each identifier and saves it to the LMS when logging is turned on.

The HoloRena player also has a built-in console for debugging and development. Beyond the set of default console commands, courseware developers can implement their custom console features, observe the internals of the framework, trace debug information from *SubApplications* and *Actions,* monitor the memory consumption of the running course.

## 4. APPLICATION OF HOLORENA

In this section we present a few case studies to give an impression how HoloRena can be used for the development of advanced web-based educational resources.

### 4.1 COPS: a game-based learning resource for teaching essential computer skills

COPS allows students to increase their computer operations preparatory skills prior to attending computer related face-to-face courses. Students will acquire and improve on the skills necessary to use a computer effectively and proficiently, while gaining knowledge of computer terminology.

#### 4.1.1 Target audience

This course has been designed for sworn law enforcement officers, probation/parole officers, prosecutors, and law enforcement or regulatory support staff whose duties include the investigation and prosecution of crimes in which high technology and the examination of electronic evidence are involved.

#### 4.1.2 Courseware overview

The COPS course represents a virtual training centre, called "The COPS Boot Camp". To boost the game-like user experience, learners have to choose a fantasy-name (code name) and they take the role of a secret agent, who has to successfully complete an exam mission, which is a real world-like evidence-collecting activity. To prepare to the exam, they can choose to take any of the three training missions.

In training missions the student is taken through a heterogeneous sequence of learning material, free practices and goal-based skill tests. This sequential presentation was based on pairing a game-story to the skill- and knowledge dependency trees built on the previously specified set of terminal learning objectives (TLOs). Hence each mission represents a linear story line. Hook-ups are available for jumping backwards and forward at some places. These hook-ups do not violate the logic of the story line and make the application of adaptive strategies possible. In skill-tests learners have to handle virtual objects in a simulated environment, such as a virtual desktop with different hardware accessories, or a file browser and text editor on a virtual laptop screen. Some of these tests are timed: the learner has only a specific amount of time to reach the goal of the exercise.

The exam mission is composed of skill tests similar to ones used in the training missions, and some knowledge-specific questions served as multiple-choice questionnaires. The learner gets feedback only at the very end of the exam mission, after his performance is evaluated. This feedback directs the user back to the location of the learning materials, and free practices in training missions if necessary.

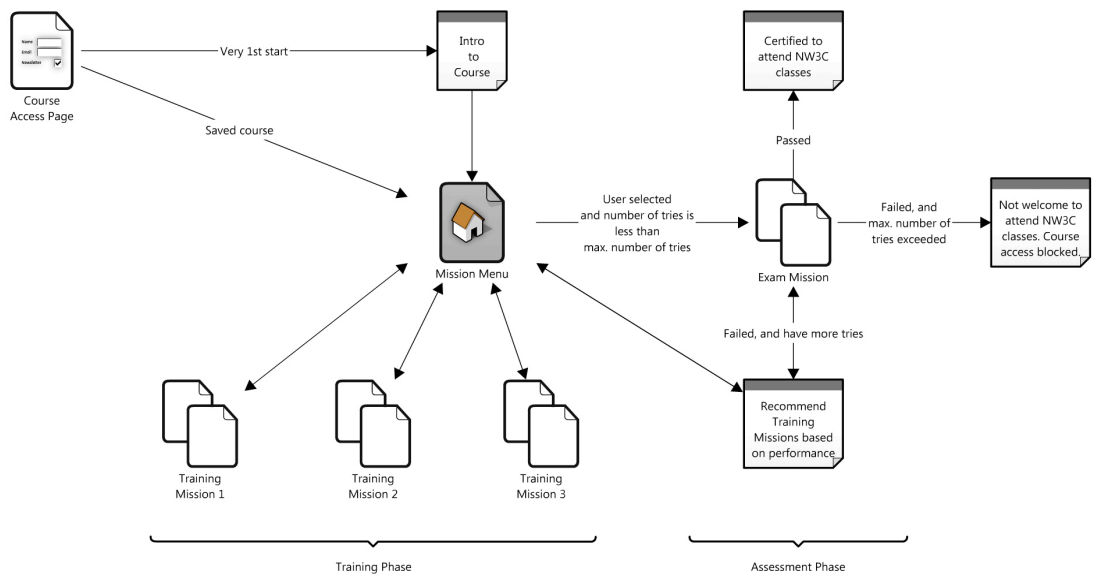

Figure 6. COPS course overview

### 4.1.3 Utilizing the HoloRena architecture for COPS

Most of the sequencing and courseware content layout is based on simple HoloRena features in COPS. The concept of *Scenes* and *Streams* made it easy and straightforward to de-couple visual representation from the textual learning content and assessment specification, and to turn story lines into a COPS course flow. Presenters were identified and designed to be used at several different places of the courseware and to work in different modes. For example, the *Hotspot Presenter* showing the back of a computer box can be used to teach about the different computer I/O ports, or it can evaluate the same knowledge in an assessment. The same *Presenter* could be used in other virtual environments, which rely on a static background image and hotspots on the screen.

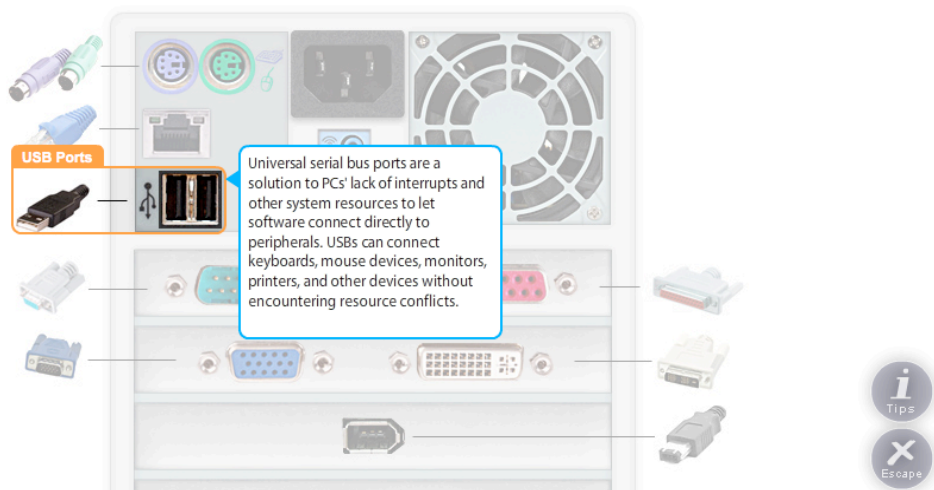

Figure 7. The Hotspot *Presenter* configured up for the "Back of a PC" tutorial

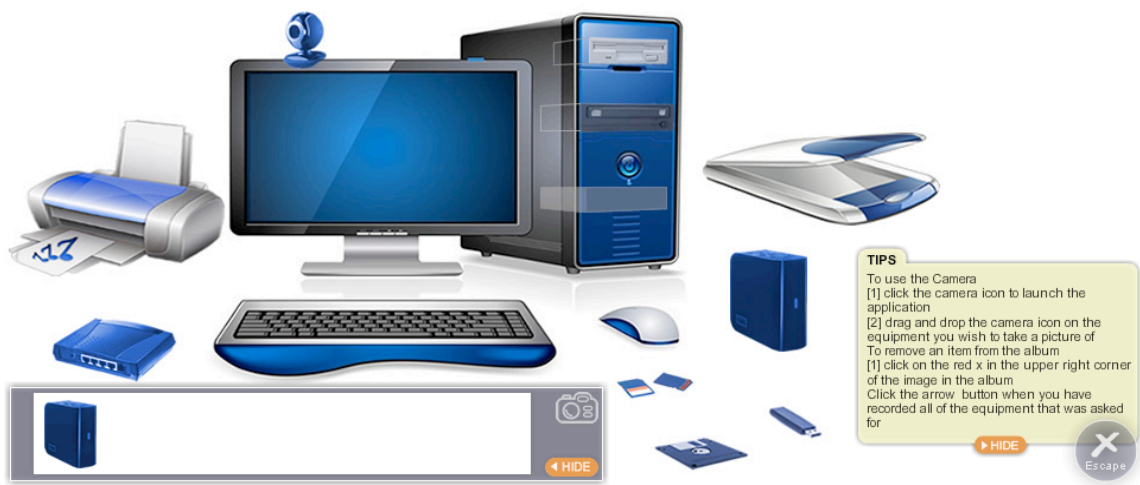

Figure 8. The Hotspot *Presenter* configured up to collaborate with the Photo Album *Gadget*

A similar example is the *Presenter,* which simulates software tools on a computer screen. It can be utilized both in tutorial mode to teach basic file operations, and in assessment mode to track user actions. The about and wrap-up screens for missions use the same type of presenters as well.

Adaptive strategies, user performance and assessment mechanism were programmed in *Action* blocks. *Gadgets* were used for cross-scene, continuous objects (e.g. timer) and to add global navigational features. For example the *Toolbar Gadget* providing access to context-aware help and to the mission menu is layered over the top of the *Presenters* in all mission *Scenes*.

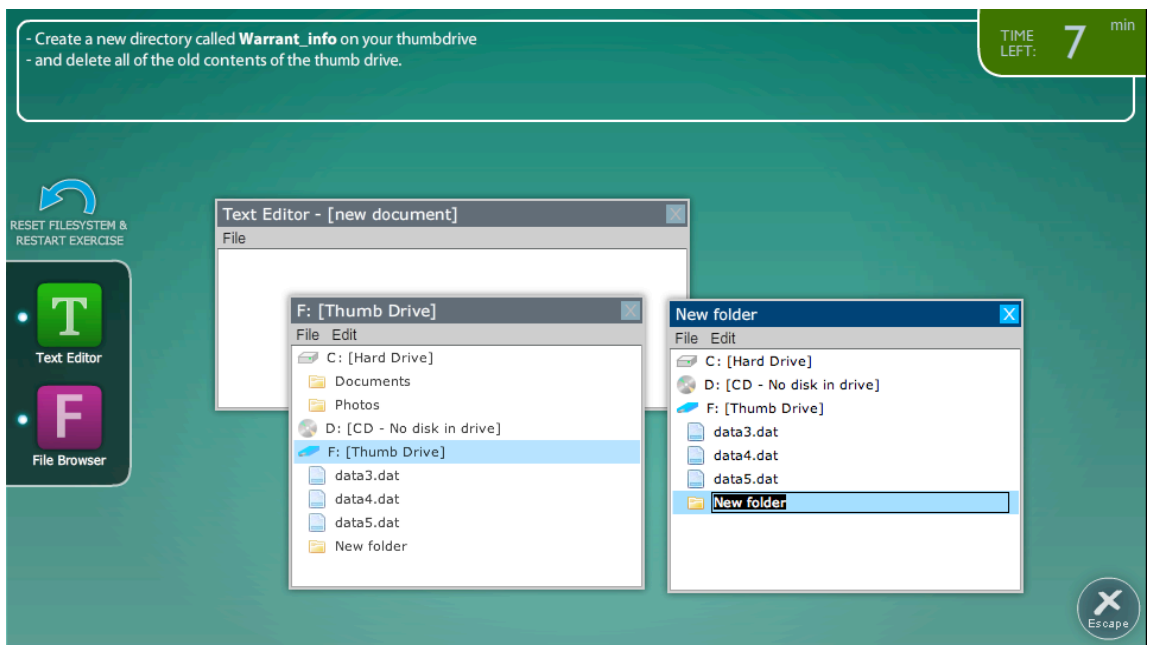

Figure 9. Virtual Desktop, with the Escape button and the timer from the ToolBar *Gadget*

HoloRena supports collaboration and distributed development: *Presenters* and *Gadgets* were implemented separately from the rest of the course content. While programmers were working on coding and designing *Presenters* (some in Flash some in Flex, HoloRena's multi-versioned application framework supports both), the instructional designers could work on the course flow XMLs and the text-based instructional content independently. In the final version of COPS 2 *Gadgets* had to be developed and 12 different *Presenters* were used by about 100 *Scenes*.

Pilot testing sessions were organized to evaluate COPS as a learning resource. The HoloRena journaling feature turned out to be very useful to analyze user interactions, identify bugs or other flaws in the user experience through mining the data collected from test-deliveries.

## 4.2 PERPS: knowledge diagnosis and remediation with scenario-based learning

This web based pre-assessment was designed to provide learners with knowledge and skill sets needed to enhance their comprehension of material covered in the face-to-face classroom during cyber forensics courses and to reduce the amount of remedial assistance provided by instructors.

### 4.2.1 Resource structure

The high-level design of PERPS centers on a prerequisite inventory (PI) for the face-to-face training. Individual items from this inventory provide the basis for the diagnostic and remedial units comprising the design. The organization of individual PI items (PII) is influenced by strategies employed for diagnosis or remediation.

The strategy for diagnosing performance deficiencies is based on scenario-based assessment: a question stream situated within a domain-specific narrative. This strategy addresses the need for authenticity and engagement by posing questions within situations related to the participant's workplace

Each diagnostic assessment in PERPS is comprised of several scenarios. A scenario incorporates typically one Diagnostic Unit (DU): sets of questions that take advantage of the context provided by the scenario to test a learner's knowledge of certain PIIs.

Remediation is provided upon completion of the initial diagnosis in the form of displaying corresponding in-place, or external remediation resources. In-place remediation resources are displayed as one-page slides, within the courseware user interface. External remediation resources are downloadable files or links to websites.

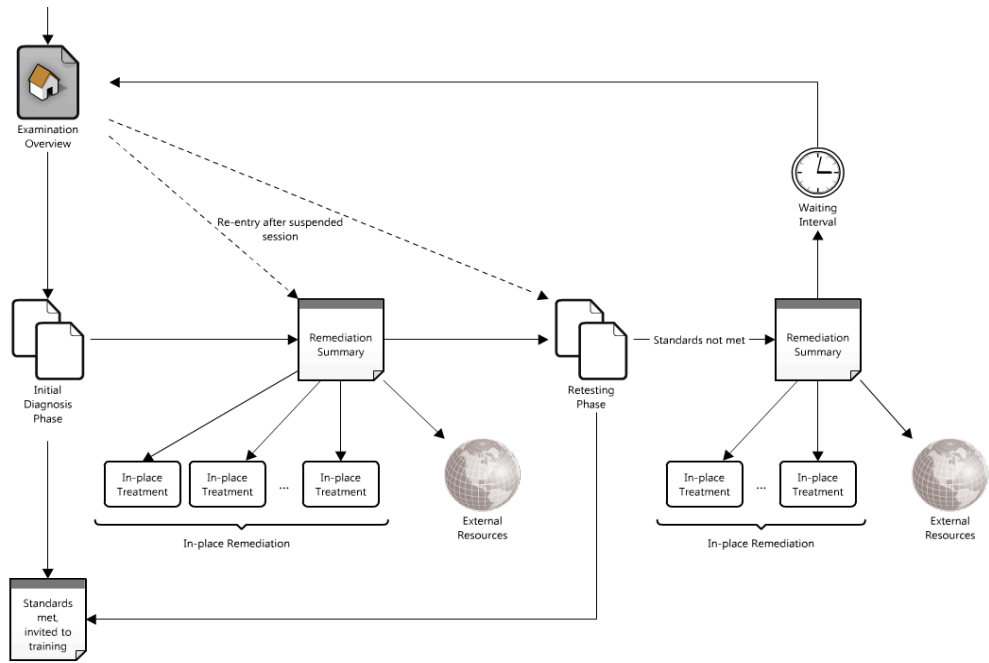

Figure 10. PERPS course structure

### 4.2.2 Generating HoloRena course flows for PERPS

To provide unique and adequate sequence of DUs during the initial test and the re-test for each learner, dynamic *Stream* content generation was excessively utilized for the implementations of the diagnostic sessions in PERPS.

Since DUs represent a static scenario instance structured as an introduction, various number questions, and a wrap-up, first coherent DU *Streams* were rendered from the content specification. Each DU *Stream* consists of an about *Scene,* followed by the sequence of question *Scenes* and a single wrap-up *Scene*. The diagnostic session *Streams* are composed of the pre-rendered DU *Streams* on the fly. The composing algorithm, called the scheduler, first determines which DU should be used to cover the PIIs the learner should be tested on. Then the *Streams* of these DUs are generated in a random order into a blank diagnostic session placeholder *Stream*.

The scheduler, the data representing the mapping of DUs to PIIs, user profiling, assessment logic and other utility functions are implemented as externalized *Action* contents.

Data specification of PERSP courses is stored in a relational database. The content authors have a from-based interface to enter and maintain the content of the scenarios and questions, absolutely independently from the presentation and the courseware-logic layer. The HoloRena course flow specification is generated directly from this back-end infrastructure.

### 4.2.3 *SubApplications* in PERPS

Though there are several hundred question *Scenes* in the PERPS course flow, only a few presenters had to be created:

- DU about and wrap-up *Presenters*.
- A *Presenter* for each question type and layout.
- A course map *Presenter*.

Using Flash based technology was a big advantage, especially in the development of interactive advanced *Presenters* for some of the questions. We could reuse the Toolbar *Gadget* we built for the COPS course.

Figure 11. The drag-and-drop PERPS *Presenter*

The same question *Presenters* can be used in playback mode during remediation, when they are layered over a *Gadget*, implementing the remediation features of the courseware. *Presenters* in this mode display feedback on the answers they recorded in diagnostic mode.

Figure 12. Fill in the blank *Presenter* in playback mode, encapsulated in the remediation *Gadget*

## 4.3 Concept of an interactive learning module with complex visual representation

Students in graphically represented exercises typically interact with visual elements displayed on a screen that has limited dimensions. This limitation can vary for different target devices and audiences. During the planning of these activities, and development of frameworks and tools which facilitate the authoring of such activities, as a general first step, designers specify the target screen size (or a set of screen sizes) for the product and then try to layout their concepts in the form of user interface (UI) plans fitting in them. In some cases, presentation of a problem on a single screen framed to an inevitably limited screen size would result in busy, overcomplicated visual representation. This could make the recognition and interpretation of the visual information, and interaction with the UI overwhelming, and hard for the student.

An example for such a situation is the digital representation of House of Quality (HoQ) diagrams. While HoQ on paper is a convenient tool to work with, when displayed on a single general-sized computer screen, it can be confusing and hard to handle. Especially when instructions and other UI controls need to be displayed on the same screen.

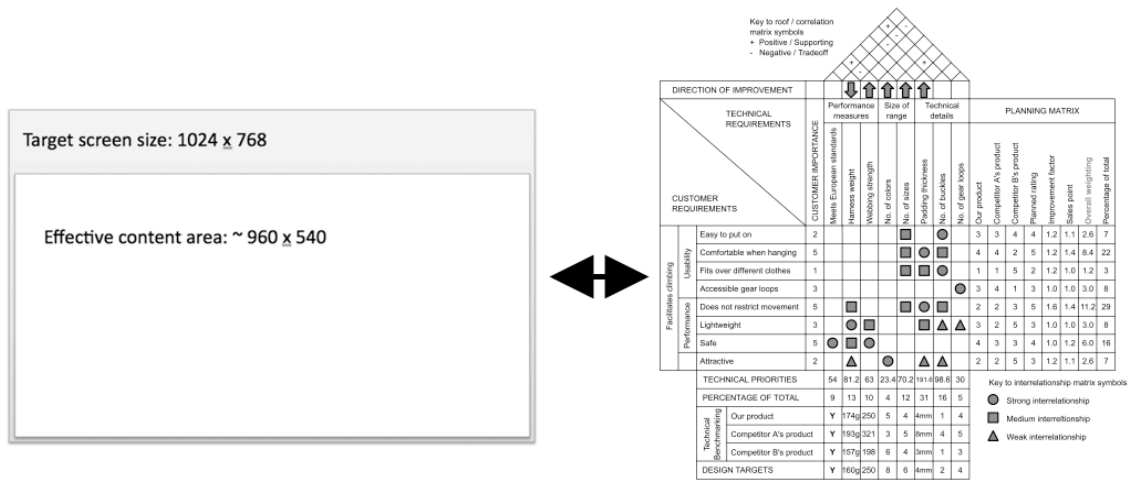

Figure 13. Screen size conflicting complexity of a HoQ exercise

We realized, that in these cases the visual representation could be divided into carefully HoloRena *Scenes*. The students would see and interact with only one segment on the screen at a time, and would be able to navigate between the *Scenes* with the use of a navigator *Gadget*. Segmentation leads to a more efficient and clearer visual layout, and hence building different parts of the house require different skills and user interactions, the need of giving only one type of problem to solve at a time to the user moderates the complexity of UI interface, too. Furthermore, these tasks quite well represent the steps of a general, complex HoQ completion process. Course designs using HoloRena in these exercises are currently under development.

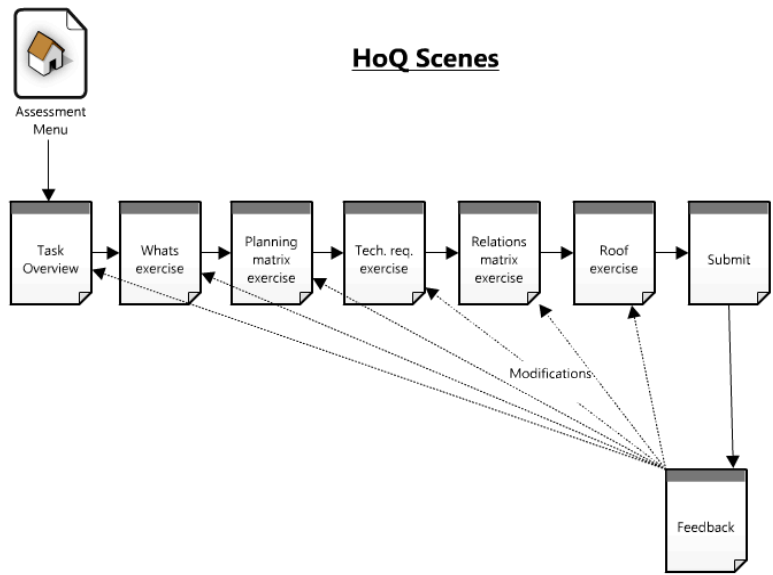

Figure 14. HoQ course structure

## 5. CONCLUSIONS AND FUTURE WORK

We would like to see more enhanced educational resources on the web. Not just because we believe that people like to work with innovative user interfaces and games and it is a huge factor in the acceptance of computer-based learning, furthermore sophisticated instructional design is needed to effectively target wide audience, but also because we think that the current technology provides much more possibilities than what it is utilized for.

Based on our past and current experiments, HoloRena is proven to be a practical framework for delivering interactive, scalable advanced educational applications on the web. However, creating such a framework was a significant amount of work, the invested effort paid off during courseware development. We had more flexibility in adaption logic and courseware design, more freedom and time to focus on interactive features, changing and fine-tuning the course content was very simple and we virtually did not have to deal with browser- or platform-compatibility issues. Educational resources built on HoloRena are entirely executed on the client-side. Developers do not have to implement complicated courseware features on the LMS, therefore HoloRena courses can very easily made to be LMS-agnostic and SCORM compliant without serious compromises.

Due to continuous improvements and tunings during experiments, HoloReana has reached a stable, release-state. We are looking for new opportunities where it can be utilized. It would be interesting to see how it could help to implement real-time collaborative exercises situated in online virtual environments. Another exciting use-case would be a multimedia-enhanced embedded training, or mobile-learning solution deployed to Flash-enabled devices.

HoloRena is a low-level framework. However it simplifies and helps the courseware development process and can serve as a strong base for GBL and SBL educational resources it is not a complete authoring environment. HoloRena course flows do not necessary derive and keep all the aspects of the high-level courseware design. Thus we are planning to add HoloRena supporting features to the CAPE authoring tool [12].

# Authors

**Laszlo Juracz**
*Human-Computer Interaction Specialist*

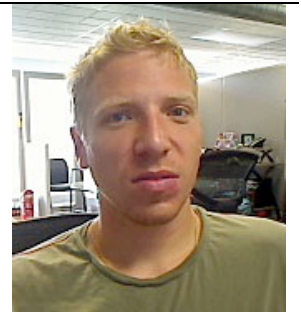

I remember the big excitement I felt when I started playing around with creating tiny visual and audio programs on a C16 in middle school. That was in the 80's. Since then computers and applications came along - a long way, multimedia and the Web were born, digital art is around us everywhere, every moment, in plenty of different forms… Although my primary motivation has evolved, as well, it has not changed that much: I think that one of the most interesting and challenging thing in my work and a basic principle in life is sharing information, message or knowledge with others and making services accessible in the most effective, convenient and entertaining way. I received my Master's degree in computer science from Budapest University of Technology and Economics in 2002 and I joined the Prototus Project at Vanderbilt in 2007.